\documentclass{article}

\usepackage{latexsym,amssymb,textcomp}

\def\AA{\mathcal A}  \def\CC{\mathcal C}
 \def\FF{\mathcal F} 
  
  \def\UU{\mathcal U}

\def\NNN{\mathbb N} \def\RRR{\mathbb R} 
  \def\BBB{\mathbb B}

\def\plt{<^{{}^{\!\!\!\!\!\!\!\;+}}}   
\def\pgt{>^{{}^{\!\!\!\!\!\!\!\;+}}}    
\def\peq{\stackrel{\scriptscriptstyle{+}}{=}}     

\newtheorem{nummer}{\hspace*{-0.33em}}[section]
\newenvironment{definition} {\begin{nummer}{\bf Definition.} \begin{rm}}{\end{rm} \end{nummer}}
\newenvironment{lemma}      {\begin{nummer} {\bf Lemma.}}       {\end{nummer}}
\newenvironment{theorem}    {\begin{nummer} {\bf Theorem.}}     {\end{nummer}}

\newenvironment{proof}      {\noindent \bf Proof. \rm} {\ \nolinebreak \hfill $\Box$ \vspace{2ex}}

\title{Is there an Elegant Universal\\ Theory of Prediction?
\thanks{This work was
supported by SNF grant 200020-107616.}}

\author{Shane Legg\thanks{\tt shane@idsia.ch}}

\begin{document}

\maketitle

\begin{abstract}
Solomonoff's inductive learning model is a powerful, universal and
highly elegant theory of sequence prediction.  Its critical flaw is
that it is incomputable and thus cannot be used in practice.  It is
sometimes suggested that it may still be useful to help guide the
development of very general and powerful theories of prediction which
are computable.  In this paper it is shown that although powerful
algorithms exist, they are necessarily highly complex.  This alone
makes their theoretical analysis problematic, however it is further
shown that beyond a moderate level of complexity the analysis runs
into the deeper problem of G\"{o}del incompleteness.  This limits the
power of mathematics to analyse and study prediction algorithms, and
indeed intelligent systems in general.
\end{abstract}

\section{Introduction}

Could there exist an elegant and universal theory of sequence
prediction?  Solomonoff's model of induction rapidly learns to make
optimal predictions for any computable sequence, including
probabilistic ones \cite{Solomonoff:64,Solomonoff:78}.  Indeed the
problem of sequence prediction could well be considered solved
\cite{hutter:06usp,hutter:04uaibook}, if it were not for the fact that
Solomonoff's theoretical model is incomputable.

Among computable theories there exist powerful general predictors,
such as the Lempel-Ziv algorithm \cite{Feder:92} and Context Tree
Weighting \cite{Willems:95}, that can learn to predict some complex
sequences, but not others.  Some prediction methods, such as the
Minimum Description Length principle \cite{Rissanen:96} and the
Minimum Message Length principle \cite{Wallace:68}, can even be viewed
as computable approximations to Solomonoff induction~\cite{Li:97}.
However in practice their power and generality are limited by the
power of compression and coding methods employed, as well as having a
significantly reduced data efficiency as compared to Solomonoff
induction \cite{Poland:04mdl2p}.

Could there exist elegant computable prediction algorithms that are in
some sense universal, or at least universal over large sets of simple
sequences?  In this paper we explore this fundamental question from
the perspective of Kolmogorov complexity theory and uncover some
surprising implications.

\section{Preliminaries}

An \emph{alphabet} $\AA$ is a finite set of 2 or more elements which
are called \emph{symbols}.  In this paper we will assume a binary
alphabet $\BBB := \{ 0, 1 \}$, though all the results can easily be
generalised to other alphabets.  A \emph{string} is a finite ordered
$n$-tuple of symbols denoted $x := x_1 x_2 \ldots x_n$ where $\forall
i \in \{ 1, \ldots, n \}$, $x_i \in \BBB$, or more succinctly, $x \in
\BBB^n$.  The 0-tuple is denoted $\lambda$ and is called the
\emph{null string}.  The expression $\BBB^{\leq n}$ has the obvious
interpretation, and $\BBB^* := \bigcup_{n \in \NNN} \BBB^n$.  The
length \emph{lexicographical} ordering is a total order on $\BBB^*$
defined as $\lambda < 0 < 1 < 00 < 01 < 10 < 11 < 000 < 001 < \cdots$.
A \emph{substring} of $x$ is defined $x_{j:k} := x_j x_{j+1} \ldots
x_k$ where $1 \leq j \leq k \leq n$.  By $|x|$ we mean the length of
the string $x$, for example, $|x_{j:k}| = k - j +1$.  We will
sometimes need to encode a natural number as a string.  Using simple
encoding techniques it can be shown that there exists a computable
injective function $f : \NNN \to \BBB^*$ where no string in the range
of $f$ is a prefix of any other, and $\forall n \in \NNN : |f(n)| \leq
\log_2 n + 2 \log_2 \log_2 n + 1$.

Unlike strings which always have finite length, a \emph{sequence}
$\omega$ is an infinite list of symbols $x_1 x_2 x_3 \ldots \in
\BBB^\infty$.  Of particular interest to us will be the class of
sequences which can be generated by an algorithm executed on a
universal Turing machine:

\begin{definition}
A {\bf monotone universal Turing machine} $\UU$ is defined as a
universal Turing machine with one unidirectional input tape, one
unidirectional output tape, and some bidirectional work tapes.  Input
tapes are read only, output tapes are write only, unidirectional tapes
are those where the head can only move from left to right.  All tapes
are binary (no blank symbol) and the work tapes are initially filled
with zeros.  We say that $\UU$ outputs/computes a sequence $\omega$ on
input $p$, and write $\UU(p) = \omega$, if $\UU$ reads all of $p$ but
no more as it continues to write $\omega$ to the output tape.
\end{definition}

We fix $\UU$ and define $\UU( p, x )$ by simply using a standard
coding technique to encode a program $p$ along with a string $x \in
\BBB^*$ as a single input string for $\UU$.

\begin{definition}
A sequence $\omega \in \BBB^\infty$ is a {\bf computable binary
sequence} if there exists a program $q \in \BBB^*$ that writes
$\omega$ to a one-way output tape when run on a monotone universal
Turing machine $\mathcal{U}$, that is, $\exists q \in \BBB^* : \UU(q)
= \omega$. We denote the set of all computable sequences by $\CC$.
\end{definition}

A similar definition for strings is not necessary as all strings have
finite length and are therefore trivially computable.

\begin{definition}
A {\bf computable binary predictor} is a program $p \in \BBB^*$ that
on a universal Turing machine $\UU$ computes a total function $\BBB^*
\to \BBB$.
\end{definition}

For simplicity of notation we will often write $p(x)$ to mean the
function computed by the program $p$ when executed on $\UU$ along with
the input string $x$, that is, $p(x)$ is short hand for $\UU( p, x )$.
Having $x_{1:n}$ as input, the objective of a predictor is for its
output, called its \emph{prediction}, to match the next symbol in the
sequence.  Formally we express this by writing $p(x_{1:n}) = x_{n+1}$.

As the algorithmic prediction of incomputable sequences, such as the
halting sequence, is impossible by definition, we only consider the
problem of predicting computable sequences.  To simplify things we
will assume that the predictor has an unlimited supply of computation
time and storage.  We will also make the assumption that the predictor
has unlimited data to learn from, that is, we are only concerned with
whether or not a predictor can learn to predict in the following
sense:

\begin{definition}
We say that a predictor $p$ can {\bf learn to predict} a sequence
$\omega := x_1 x_2 \ldots \in \BBB^\infty$ if there exists $m \in \NNN$
such that $\forall n \geq m : p(x_{1:n}) = x_{n+1}$.
\end{definition}

The existence of $m$ in the above definition need not be constructive,
that is, we might not know when the predictor will stop making
prediction errors for a given sequence, just that this will occur
eventually.  This is essentially ``next value'' prediction as
characterised by Barzdin~\cite{Barzdin:72}, which follows from Gold's
notion of identifiability in the limit for languages~\cite{Gold:67}.

\begin{definition}
Let $P(\omega)$ be the set of all predictors able to learn to predict
$\omega$.  Similarly for sets of sequences $S \subset
\BBB^\infty$, define $P(S) := \bigcap_{\omega \in S} P( \omega )$.
\end{definition}

A standard measure of complexity for sequences is the length of the
shortest program which generates the sequence:
\begin{definition}
For any sequence $\omega \in \BBB^\infty$ the monotone {\bf Kolmogorov
complexity} of the sequence is,
\[
K( \omega ) := \min_{q \in \BBB^*} \{ |q| : \UU(q) = \omega \},
\]
where $\UU$ is a monotone universal Turing machine.  If no such $q$
exists, we define $K(\omega) := \infty$.
\end{definition}

It can be shown that this measure of complexity depends on our choice
of universal Turing machine $\UU$, but only up to an additive constant
that is independent of $\omega$.  This is due to the fact that a
universal Turing machine can simulate any other universal Turing
machine with a fixed length program.

In essentially the same way as the definition above we can define the
Kolmogorov complexity of a string $x \in \BBB^n$, written $K(x)$, by
requiring that $\UU(q)$ halts after generating $x$ on the output tape.
For an extensive treatment of Kolmogorov complexity and some of its
applications see \cite{Li:97} or \cite{Calude:02}.

As many of our results will have the above property of holding within
an additive constant that is independent of the variables in the
expression, we will indicate this by placing a small plus above the
equality or inequality symbol.  For example, $f(x) \plt g(x)$ means
that that $\exists c \in \RRR, \forall x : f(x) < g(x) + c$.  When
using standard ``Big O'' notation this is unnecessary as expressions
are already understood to hold within an independent constant, however
for consistency of notation we will use it in these cases also.

\section{Prediction of computable sequences}

The most elementary result is that every computable sequence can be
predicted by at least one predictor, and that this predictor need not
be significantly more complex than the sequence to be predicted.

\begin{lemma}\label{lem:bound1}
$\forall \omega \in \CC, \exists p \in P( \omega ) : K( p )
\plt K( \omega )$.
\end{lemma}

\begin{proof}
As the sequence $\omega$ is computable, there must exist at least one
algorithm that generates $\omega$.  Let $q$ be the shortest such
algorithm and construct an algorithm $p$ that ``predicts'' $\omega$ as
follows: Firstly the algorithm $p$ reads $x_{1:n}$ to find the value
of $n$, then it runs $q$ to generate $x_{1:n+1}$ and returns $x_{n+1}$
as its prediction.  Clearly $p$ perfectly predicts $\omega$ and $|p| <
|q| + c$, for some small constant $c$ that is independent of $\omega$
and $q$. 
\end{proof}

Not only can any computable sequence be predicted, there also exist
very simple predictors able to predict arbitrarily complex sequences:

\begin{lemma}\label{lem:predofcomplex}
There exist a predictor $p$ such that $\forall n \in \NNN, \exists \,
\omega \in \CC : p \in P(\omega)$ and $K(\omega) > n$.
\end{lemma}

\begin{proof}
Take a string $x$ such that $K(x) = |x| \geq 2n$, and from this
define a sequence $\omega := x 0 0 0 0 \ldots$.  Clearly $K(\omega) >
n$ and yet a simple predictor $p$ that always predicts 0 can learn to
predict $\omega$. 
\end{proof}

The predictor used in the above proof is very simple and can only
learn sequences that end with all 0's, albeit where the initial string
can have an arbitrarily high Kolmogorov complexity.  It is not hard to
see that more sophisticated predictors can learn to predict many other
more subtle types of patterns which are more complex than the
predictor, such as arbitrary repeating strings, regular or primitive
recursive sequences.

As each computable sequence can be predicted, and simple predictors
exist which can predict arbitrarily complex sequences, we might wonder
whether there exists a computable predictor able to learn to predict
all computable sequences.  Unfortunately, no universal predictor
exists, indeed for every predictor there exists a sequence which it
cannot predict at all:

\begin{lemma}\label{lem:adv}
For any predictor $p$ there constructively exists a sequence $\omega
:= x_1 x_2 \ldots \in \CC$ such that $\forall n \in \NNN : p(x_{1:n})
\neq x_{n+1}$ and $K(\omega) \plt K(p)$.
\end{lemma}

\begin{proof}
For any computable predictor $p$ there constructively exists a
computable sequence $\omega = x_1 x_2 x_3 \ldots$ computed by an
algorithm $q$ defined as follows: Set $x_1 = 1 - p(\lambda)$, then
$x_2 = 1 - p( x_1 )$, then $x_3 = 1 - p( x_{1:2} )$ and so on.
Clearly $\omega \in \CC$ and $\forall n \in \NNN : p(x_{1:n}) = 1 -
x_{n+1}$.

Let $p^*$ be the shortest program that computes the same function as
$p$ and define a sequence generation algorithm $q^*$ based on $p^*$
using the procedure above.  By construction, $|q^*| = |p^*| + c$ for
some constant $c$ that is independent of $p^*$.  Because $q^*$
generates $\omega$, it follows that $K(\omega) \leq |q^*|$.  By
definition $K(p) = |p^*|$ and so $K(\omega) \plt K(p)$.
\end{proof}

Allowing the predictor to be probabilistic does not fundamentally
avoid the problem of Lemma~\ref{lem:adv}.  In each step, rather than
generating the opposite to what will be predicted by $p$, instead $q$
attempts to generate the symbol which $p$ is least likely to predict
given $x_{1:n}$.  To do this $q$ must simulate $p$ in order to
estimate $\Pr \! \big( p(x_{1:n}) = 1 \big| x_{1:n} \big)$.  With
sufficient simulation effort, $q$ can estimate this probability to any
desired accuracy for any $x_{1:n}$.  This produces a computable
sequence $\omega$ such that $\forall n \in \NNN : \Pr \!  \big(
p(x_{1:n}) = x_{n+1} \big| x_{1:n} \big)$ is not significantly greater
than $\frac{1}{2}$, that is, the performance of $p$ is no better than
a predictor that makes completely random predictions.  

The impossibility of prediction in this more general probabilistic
setting has been pointed out before by Dawid~\cite{Dawid:85}.
Specifically, Dawid notes that for any statistical forecasting system
there exist sequences which are not calibrated.  Dawid also notes that
a forecasting system for a family of distributions is necessarily more
complex than any forecasting system generated from a single
distribution in the family.  However, he does not deal with the
complexity of the sequences themselves, nor does he make a precise
statement in terms of a specific measure of complexity, such as
Kolmogorov complexity.  The impossibility of forecasting has since
been developed in considerably more depth by V'yugin~\cite{Vyugin:98},
in particular, it is proven that there is an efficient randomised
procedure producing sequences that cannot be predicted (with high
probability) by computable forecasting systems.

As probabilistic prediction complicates things without avoiding this
fundamental problem, in the remainder of this paper we will consider
only deterministic predictors.  This will also allow us to see the
roots of this problem as clearly as possible.  With the preliminaries
covered, we now move on to the central problem considered in this
paper: Predicting sequences of limited Kolmogorov complexity.

\section{Prediction of simple computable sequences}

As the computable prediction of any computable sequence is impossible,
a weaker goal is to be able to predict all ``simple'' computable
sequences.

\begin{definition}
For $n \in \NNN$, let $\CC_n := \{ \omega \in \CC: K(\omega) \leq n
\}$.  Further, let $P_n := P( \CC_n )$ be the set of predictors able
to learn to predict all sequences in $\CC_n$.
\end{definition}

Firstly we establish that prediction algorithms exist that can learn
to predict all sequences up to a given complexity, and that these
predictors need not be significantly more complex than the sequences
they can predict:

\begin{lemma} \label{lem:infpredictors}
$\forall n \in \NNN, \exists p \in P_n : K( p ) \plt n + O( \log_2 n )$.
\end{lemma}

\begin{proof}
Let $h \in \NNN$ be the number of programs of length $n$ or less which
generate infinite sequences.  Build the value of $h$ into a prediction
algorithm $p$ constructed as follows:

In the $k^{th}$ prediction cycle run in parallel all programs of
length $n$ or less until $h$ of these programs have each produced
$k+1$ symbols of output.  Next predict according to the $k+1^{th}$
symbol of the generated string whose first $k$ symbols is consistent
with the observed string.  If two generated strings are consistent
with the observed sequence (there cannot be more than two as the
strings are binary and have length $k+1$), pick the one which was
generated by the program that occurs first in a lexicographical
ordering of the programs.  If no generated output is consistent, give
up and output a fixed symbol.

For sufficiently large $k$, only the $h$ programs which produce
infinite sequences will produce output strings of length $k$.  As this
set of sequences is finite, they can be uniquely identified by finite
initial strings.  Thus for sufficiently large $k$ the predictor $p$
will correctly predict any computable sequence $\omega$ for which $K(
\omega ) \leq n$, that is, $p \in P_n$.

As there are $2^{n+1} -1$ possible strings of length $n$ or less, $h <
2^{n+1}$ and thus we can encode $h$ with $\log_2 h + 2 \log_2 \log_2 h
= n + 1 + 2\log_2 (n+1)$ bits.  Thus, $K( p ) < n + 1 + 2 \log_2 (n+1)
+ c$ for some constant $c$ that is independent of $n$. 
\end{proof}

Can we do better than this?  Lemma~\ref{lem:predofcomplex} shows us
that there exist predictors able to predict at least some sequences
vastly more complex than themselves.  This suggests that there might
exist simple predictors able to predict arbitrary sequences up to a
high complexity.  Formally, could there exist $p \in P_n$ where $n \gg
K(p)$?  Unfortunately, these simple but powerful predictors are not
possible:

\begin{theorem}\label{thm:simplepred}
$\forall n \in \NNN: p \in P_n \Rightarrow K(p) \pgt n$.
\end{theorem}

\begin{proof}
For any $n \in \NNN$ let $p \in P_n$, that is, $\forall \omega \in
\CC_n: p \in P(\omega)$.  By Lemma~\ref{lem:adv} we know that $\exists
\, \omega' \in \CC : p \notin P(\omega')$ .  As $p \notin P(\omega')$
it must be the case that $\omega' \notin \CC_n$, that is, $K(\omega')
\geq n$.  From Lemma~\ref{lem:adv} we also know that $K(p) \pgt
K(\omega')$ and so the result follows.  
\end{proof}

Intuitively the reason for this is as follows: Lemma~\ref{lem:adv}
guarantees that every simple predictor fails for at least one simple
sequence.  Thus if we want a predictor that can learn to predict all
sequences up to a moderate level of complexity, then clearly the
predictor cannot be simple.  Likewise, if we want a predictor that can
predict all sequences up to a high level of complexity, then the
predictor itself must be very complex.  Thus, even though we have made
the generous assumption of unlimited computational resources and data
to learn from, only very complex algorithms can be truly powerful
predictors.

These results easily generalise to notions of complexity that take
computation time into consideration.  As sequences are infinite, the
appropriate measure of time is the time needed to generate or predict
the next symbol in the sequence.  Under any reasonable measure of time
complexity, the operation of inverting a single output from a binary
valued function can be performed with little cost.  If $C$ is any
complexity measure with this property, it is trivial to see that the
proof of Lemma~\ref{lem:adv} still holds for $C$.  From this, an
analogue of Theorem~\ref{thm:simplepred} for $C$ easily follows.  With
similar arguments these results also generalise in a straightforward
way to complexity measures that take space or other computational
resources into account.  Thus, the fact that extremely powerful
predictors must be very complex, holds under any measure of complexity
for which inverting a single bit is inexpensive.

\section{Complexity of prediction}

Another way of viewing these results is in terms of an alternate
notion of sequence complexity defined as the size of the smallest
predictor able to learn to predict the sequence.  This allows us to
express the results of the previous sections more concisely.
Formally, for any sequence $\omega$ define the complexity measure,
\[
\dot{K} ( \omega ) := \min_{p \in \BBB^*} \{ |p| : p \in P( \omega ) \},
\]
and $\dot{K}(\omega) := \infty$ if $P( \omega ) = \varnothing$.  Thus,
if $\dot{K} ( \omega )$ is high then the sequence $\omega$ is complex
in the sense that only complex prediction algorithms are able to learn
to predict it.  It can easily be seen that this notion of complexity
has the same invariance to the choice of reference universal Turing
machine as the standard Kolmogorov complexity measure.

It may be tempting to conjecture that this definition simply describes
what might be called the ``tail end complexity'' of a sequence, that
is, $\dot{K}(\omega) = \lim_{i \to \infty} K(\omega_{i:\infty})$.
This is not the case.  Consider again Lemma~\ref{lem:predofcomplex}
and its proof.  For any $n \in \NNN$, we let $y_{1:n}$ be a random
string, that is, $K(y_{1:n}) \peq n$.  From this we defined a
computable sequence that was a repetition of this string, $\omega :=
(y_{1:n})^*$.  It was then proven that there exists a single predictor
$p$ which can predict any sequence of this form, with no restriction
on how high $K(\omega)$ can be.  From our definition of $\dot{K}$
above it is thus clear that $\dot{K}(\omega) \peq 0$ for any such
$\omega$.  Consider now the tail complexity of $\omega$.  As
$K(y_{1:n}) \peq n$, whenever $i \bmod n = 0$ we have
$K(\omega_{i:\infty}) \pgt n - O(\log n)$ (the $O(\log n)$ term comes
from potentially saving bits due to not having to encode $|y_{1:n}|$).
Thus even if the limit $\lim_{i \to \infty} K(\omega_{i:\infty})$
exists (it may oscillate), it cannot be equal to $\dot{K}(\omega)$ in
general.

Using $\dot{K}$ we can now rewrite a number of our previous results
more succinctly in terms of the new complexity measure.  From
Lemma~\ref{lem:bound1} it immediately follows that,
\[
\forall
\omega: 0 \leq \dot{K}( \omega ) \plt K( \omega ).
\]
From Lemma~\ref{lem:predofcomplex} we know that $\exists c \in \NNN,
\forall n \in \NNN, \exists \, \omega \in \CC$ such that $\dot{K}(
\omega ) <c$ and $K( \omega ) > n$, that is, $\dot{K}$ can attain the
lower bound above within a small constant, no matter how large the
value of $K$ is.  The sequences for which the upper bound on $\dot{K}$
is tight are interesting as they are the ones which demand complex
predictors.  We prove the existence of these sequences and look at
some of their properties in the next section.

The complexity measure $\dot{K}$ can also be generalised to sets of
sequences, for $S \subset \BBB^\infty$ define $\dot{K}( S ) := \min_p
\{ |p|: p \in P(S) \}$.  This allows us to rewrite
Lemma~\ref{lem:infpredictors} and Theorem~\ref{thm:simplepred} as
simply,
\[
\forall n \in \NNN : n
\plt \dot{K} ( \CC_n ) \plt n + O( \log_2 n ).
\]
This is just a restatement of the fact that the simplest predictor
capable of predicting all sequences up to a Kolmogorov complexity of
$n$, has itself a Kolmogorov complexity of roughly $n$.

\section{Hard to predict sequences}\label{sec:hard}

We have already seen that some individual sequences, such as the
repeating string used in the proof of Lemma~\ref{lem:predofcomplex},
can have arbitrarily high Kolmogorov complexity but nevertheless can
be predicted by trivial algorithms.  Thus, although these sequences
contain a lot of information in the Kolmogorov sense, in a deeper
sense their structure is very simple and easily learnt.

What interests us in this section is the other extreme; individual
sequences which can only be predicted by complex predictors.  As we
are only concerned with prediction in the limit, this extra complexity
in the predictor must be some kind of special information which
cannot be learnt just through observing the sequence.  Our first task
is to show that these difficult to predict sequences exist.

\begin{theorem}\label{thm:uninf}
$\forall n \in \NNN, \exists \, \omega \in \CC : n \plt \dot{K}(
\omega ) \plt K(\omega) \plt n + O( \log_2 n )$.
\end{theorem}

\begin{proof}
For any $n \in \NNN$, let $Q_n \subset \BBB^{<n}$ be the set of
programs shorter than $n$ that are predictors, and let $x_{1:k} \in
\BBB^k$ be the observed initial string from the sequence $\omega$
which is to be predicted.  Now construct a meta-predictor $\hat{p}$:

By dovetailing the computations, run in parallel every program of
length less than $n$ on every string in $\BBB^{\leq k}$.  Each time a
program is found to halt on all of these input strings, add the
program to a set of ``candidate prediction algorithms'', called
$\tilde{Q}^k_n$.  As each element of $Q_n$ is a valid predictor and
thus will halt for all input strings for any $k$, for every $n$ and
$k$ it eventually will be the case that $|\tilde{Q}^k_n| = |Q_n|$.  At
this point the simulation to approximate $Q_n$ terminates.  It is
clear that for sufficiently large values of $k$ all of the valid
predictors, and only the valid predictors, will halt with a single
symbol of output on all tested input strings.  That is, $\exists r \in
\NNN, \forall k > r : \tilde{Q}^k_n = Q_n$.

The second part of the $\hat{p}$ algorithm uses these candidate
prediction algorithms to make a prediction.  For $p \in \tilde{Q}^k_n$
define $d^k(p) := \sum_{i=1}^{k-1} |p(x_{1:i})-x_{i+1}|$.  Informally,
$d^k(p)$ is the number of prediction errors made by $p$ so far.
Compute this for all $p \in \tilde{Q}^k_n$ and then let $p^*_k \in
\tilde{Q}^k_n$ be the program with minimal $d^k(p)$.  If there is more
than one such program, break the tie by letting $p^*_k$ be the
lexicographically first of these.  Finally, $\hat{p}$ computes the
value of $p^*_k(x_{1:k})$ and then returns this as its prediction and
halts.

By Lemma~\ref{lem:adv}, there exists $\omega' \in \CC$ such that
$\hat{p}$ makes a prediction error for every $k$ when trying to
predict $\omega'$.  Thus, in each cycle at least one of the finitely
many predictors with minimal $d^k$ makes a prediction error and so
$\forall p \in Q_n: d^k(p) \to \infty$ as $k \to \infty$.  Therefore,
$\nexists p \in Q_n : p \in P(\omega')$, that is, no program of length
less than $n$ can learn to predict $\omega'$ and so $n \leq
\dot{K}(\omega')$.  Further, from Lemma~\ref{lem:bound1} we know that
$\dot{K}( \omega' ) \plt K(\omega')$, and from Lemma~\ref{lem:adv}
again, $K(\omega') \plt K(\hat{p})$.

Examining the algorithm for $\hat{p}$, we see that it contains some
fixed length program code and an encoding of $|Q_n|$, where $|Q_n| <
2^n-1$.  Thus, using a standard encoding method for integers,
$K(\hat{p}) \plt n + O( \log_2 n )$.

Chaining these together we get, $n \plt \dot{K}( \omega' ) \plt
K(\omega') \plt K(\hat{p}) \plt n + O( \log_2 n )$, which proves the
theorem. 
\end{proof}

This establishes the existence of sequences with arbitrarily high
$\dot{K}$ complexity which also have a similar level of Kolmogorov
complexity.  Next we establish a fundamental property of high
$\dot{K}$ complexity sequences: they are extremely difficult to
compute.

For an algorithm $q$ that generates $\omega \in \CC$, define $t_q(n)$
to be the number of computation steps performed by $q$ before the
$n^{th}$ symbol of $\omega$ is written to the output tape.  For
example, if $q$ is a simple algorithm that outputs the sequence
$010101\ldots$, then clearly $t_q(n) = O(n)$ and so $\omega$ can be
computed quickly.  The following theorem proves that if a sequence can
be computed in a reasonable amount of time, then the sequence must
have a low $\dot{K}$ complexity:

\begin{lemma}\label{lem:slow}
$\forall \omega \in \CC$, if $\exists q : \UU(q) = \omega$ and
$\exists r \in \NNN , \forall n > r : t_q(n) < 2^n$, then
$\dot{K}(\omega) \peq 0$.
\end{lemma}

\begin{proof}
Construct a prediction algorithm $\tilde{p}$ as follows:

On input $x_{1:n}$, run all programs of length $n$ or less, each for
$2^{n+1}$ steps.  In a set $W_n$ collect together all generated
strings which are at least $n+1$ symbols long and where the first $n$
symbols match the observed string $x_{1:n}$.  Now order the strings in
$W_n$ according to a lexicographical ordering of their generating
programs.  If $W_n = \varnothing$, then just return a prediction of 1
and halt.  If $|W_n| > 1$ then return the $n+1^{th}$ symbol from the
first sequence in the above ordering.

Assume that $\exists q : \UU(q) = \omega$ such that $\exists r \in
\NNN , \forall n > r : t_q(n) < 2^n$.  If $q$ is not unique, take $q$
to be the lexicographically first of these.  Clearly $\forall n > r$
the initial string from $\omega$ generated by $q$ will be in the set
$W_n$.  As there is no lexicographically lower program which can
generate $\omega$ within the time constraint $t_q (n) < 2^n$ for all
$n>r$, for sufficiently large $n$ the predictor $\tilde{p}$ must
converge on using $q$ for each prediction and thus $\tilde{p} \in
P(\omega)$.  As $|\tilde{p}|$ is clearly a fixed constant that is
independent of $\omega$, it follows then that $\dot{K}(\omega) <
|\tilde{p}| \peq 0$.  
\end{proof}

We could replace the $2^n$ bound in the above result with an even more
rapidly growing computable function, for example, $2^{2^n}$.  In any
case, this does not change the fundamental result that sequences which
have a high $\dot{K}$ complexity are practically impossible to
compute.  However from our theoretical perspective these sequences
present no problem as they can be predicted, albeit with immense
difficulty.

\section{The limits of mathematical analysis}

One way to interpret the results of the previous sections is in terms
of constructive theories of prediction.  Essentially, a constructive
theory of prediction $\mathcal{T}$, expressed in some sufficiently
rich formal system $\mathcal{F}$, is in effect a description of a
prediction algorithm with respect to a universal Turing machine which
implements the required parts of $\mathcal{F}$.  Thus from
Theorems~\ref{thm:simplepred} and \ref{thm:uninf} it follows that if
we want to have a predictor that can learn to predict all sequences up
to a high level of Kolmogorov complexity, or even just predict
individual sequences which have high $\dot{K}$ complexity, the
constructive theory of prediction that we base our predictor on must
be very complex.  Elegant and highly general constructive theories of
prediction simply do not exist, even if we assume unlimited
computational resources.  This is in marked contrast to Solomonoff's
highly elegant but non-constructive theory of prediction.

Naturally, highly complex theories of prediction will be very
difficult to mathematically analyse, if not practically impossible.
Thus at some point the development of very general prediction
algorithms must become mainly an experimental endeavour due to the
difficulty of working with the required theory.  Interestingly, an
even stronger result can be proven showing that beyond some point the
mathematical analysis is in fact impossible, even in theory:

\begin{theorem}\label{thm:incomplete}
In any consistent formal axiomatic system $\FF$ that is sufficiently
rich to express statements of the form ``$p \in P_n$'', there exists
$m \in \NNN$ such that for all $n > m$ and for all predictors $p \in
P_n$ the true statement ``$p \in P_n$'' cannot be proven in $\FF$.
\end{theorem}

In other words, even though we have proven that very powerful sequence
prediction algorithms exist, beyond a certain complexity it is
impossible to find any of these algorithms using mathematics.  The
proof has a similar structure to Chaitin's information theoretic proof
\cite{Chaitin:82} of G\"{o}del incompleteness theorem for formal
axiomatic systems \cite{Goedel:31}.

\begin{proof}
For each $n \in \NNN$ let $T_n$ be the set of statements expressed in
the formal system $\FF$ of the form ``$p \in P_n$'', where $p$ is
filled in with the complete description of some algorithm in each
case.  As the set of programs is denumerable, $T_n$ is also
denumerable and each element of $T_n$ has finite length.  From
Lemma~\ref{lem:infpredictors} and Theorem~\ref{thm:simplepred} it
follows that each $T_n$ contains infinitely many statements of the
form ``$p \in P_n$'' which are true.

Fix $n$ and create a search algorithm $s$ that enumerates all proofs
in the formal system $\FF$ searching for a proof of a
statement in the set $T_n$.  As the set $T_n$ is recursive, $s$ can
always recognise a proof of a statement in $T_n$.  If $s$ finds any
such proof, it outputs the corresponding program $p$ and then halts.

By way of contradiction, assume that $s$ halts, that is, a proof of a
theorem in $T_n$ is found and $p$ such that $p \in P_n$ is generated
as output.  The size of the algorithm $s$ is a constant (a description
of the formal system $\FF$ and some proof enumeration code) as well as
an $O( \log_2 n )$ term needed to describe $n$.  It follows then that
$K(p) \plt O( \log_2 n )$.  However from Theorem~\ref{thm:simplepred}
we know that $K(p) \pgt n$.  Thus, for sufficiently large $n$, we have
a contradiction and so our assumption of the existence of a proof must
be false.  That is, for sufficiently large $n$ and for all $p \in
P_n$, the true statement ``$p \in P_n$'' cannot be proven within the
formal system~$\FF$. 
\end{proof}

The exact value of $m$ depends on our choice of formal system $\FF$
and which reference machine $\UU$ we measure complexity with respect
to.  However for reasonable choices of $\FF$ and $\UU$ the value of
$m$ would be in the order of 1000.  That is, the bound $m$ is
certainly not so large as to be vacuous.

\section{Discussion}

Solomonoff induction is an elegant and extremely general model of
inductive learning.  It neatly brings together the philosophical
principles of Occam's razor, Epicurus' principle of multiple
explanations, Bayes theorem and Turing's model of universal
computation into a theoretical sequence predictor with astonishingly
powerful properties.  If theoretical models of prediction can have
such elegance and power, one cannot help but wonder whether similarly
beautiful and highly general computable theories of prediction are
also possible.

What we have shown here is that there does not exist an elegant
constructive theory of prediction for computable sequences, even if we
assume unbounded computational resources, unbounded data and learning
time, and place moderate bounds on the Kolmogorov complexity of the
sequences to be predicted.  Very powerful computable predictors are
therefore necessarily complex.  We have further shown that the source
of this problem is computable sequences which are extremely expensive
to compute.  While we have proven that very powerful prediction
algorithms which can learn to predict these sequences exist, we have
also proven that, unfortunately, mathematical analysis cannot be used
to discover these algorithms due to problems of G\"{o}del
incompleteness.

These results can be extended to more general settings, specifically
to those problems which are equivalent to, or depend on, sequence
prediction.  Consider, for example, a reinforcement learning agent
interacting with an environment \cite{Sutton:98,hutter:04uaibook}.  In
each interaction cycle the agent must choose its actions so as to
maximise the future rewards that it receives from the environment.  Of
course the agent cannot know for certain whether or not some action
will lead to rewards in the future, thus it must predict these.
Clearly, at the heart of reinforcement learning lies a prediction
problem, and so the results for computable predictors presented in
this paper also apply to computable reinforcement learners.  More
specifically, from Theorem~\ref{thm:simplepred} it follows that very
powerful computable reinforcement learners are necessarily complex,
and from Theorem~\ref{thm:incomplete} it follows that it is impossible
to discover extremely powerful reinforcement learning algorithms
mathematically.

It is reasonable to ask whether the assumptions we have made in our
model need to be changed.  If we increase the power of the predictors
further, for example by providing them with some kind of an oracle,
this would make the predictors even more unrealistic than they
currently are.  Clearly this goes against our goal of finding an
elegant, powerful and general prediction theory that is more realistic
in its assumptions than Solomonoff's incomputable model.  On the other
hand, if we weaken our assumptions about the predictors' resources to
make them more realistic, we are in effect taking a subset of our
current class of predictors.  As such, all the same limitations and
problems will still apply, as well as some new ones.

It seems then that the way forward is to further restrict the problem
space.  One possibility would be to bound the amount of computation
time needed to generate the next symbol in the sequence.  However if
we do this without restricting the predictors' resources then the
simple predictor from Lemma~\ref{lem:slow} easily learns to predict
any such sequence and thus the problem of prediction in the limit has
become trivial.  Another possibility might be to bound the memory of
the machine used to generate the sequence, however this makes the
generator a finite state machine and thus bounds its computation time,
again making the problem trivial.

Perhaps the only reasonable solution would be to add additional
restrictions to both the algorithms which generate the sequences to be
predicted, and to the predictors.  We may also want to consider not
just learnability in the limit, but also how quickly the predictor is
able to learn.  Of course we are then facing a much more difficult
analysis problem.

\subsubsection*{Acknowledgements}

I would like to thank Marcus Hutter, Alexey Chernov, Daniil Ryabko and
Laurent Orseau for useful discussions and advice during the
development of this paper.

\end{document}